\documentclass[12pt]{article}

\newcommand{\V}[1]{{\bm{\mathbf{\MakeLowercase{#1}}}}}

\newcommand{\M}[1]{{\bm{\mathbf{\MakeUppercase{#1}}}}}

\newcommand{\blind}{0}

\usepackage{graphics}
\usepackage{pstricks}
\usepackage{bm}
\usepackage{amssymb}
\usepackage{perpage} 
\MakePerPage{footnote} 
\usepackage{bbm}
\usepackage{csquotes}
\usepackage{amsmath}
\usepackage{graphicx,psfrag,epsf}
\usepackage{enumerate}
\usepackage{natbib}
\usepackage{float}

\begin{document}

\def\spacingset#1{\renewcommand{\baselinestretch}%
{#1}\small\normalsize} \spacingset{1}


\if0\blind
{
  \title{\bf A Method for Controlling Extrapolation when Visualizing and Optimizing the Prediction Profiles of Statistical and Machine Learning Models}
  \author{Jeremy Ash, \hspace{.2cm}\\
    Laura Lancaster, \hspace{.2cm} \\
    and \\
    Chris Gotwalt 
    \hspace{.2cm} \\
    JMP Division, SAS Institute}
  \maketitle
} \fi

\if1\blind
{
  \bigskip
  \bigskip
  \bigskip
  \begin{center}
    {\LARGE\bf Title}
\end{center}
  \medskip
} \fi

\bigskip
\begin{abstract}

We present a novel method for controlling extrapolation in the prediction profiler in the JMP software.  The prediction profiler is a graphical tool for exploring high dimensional prediction surfaces for statistical and machine learning models. The profiler contains interactive cross-sectional views, or profile traces, of the prediction surface of a model.  Our method helps users avoid exploring predictions that should be considered extrapolation.  It also performs optimization over a constrained factor region that avoids extrapolation using a genetic algorithm.  In simulations and real world examples, we demonstrate how optimal factor settings without constraint in the profiler are frequently extrapolated, and how extrapolation control helps avoid these solutions with invalid factor settings that may not be useful to the user.

\end{abstract}

\noindent%

\spacingset{1.45}
\section{Introduction}
\label{sec:intro}

\begin{figure}
	\centering 
	\includegraphics[width=.7\linewidth]{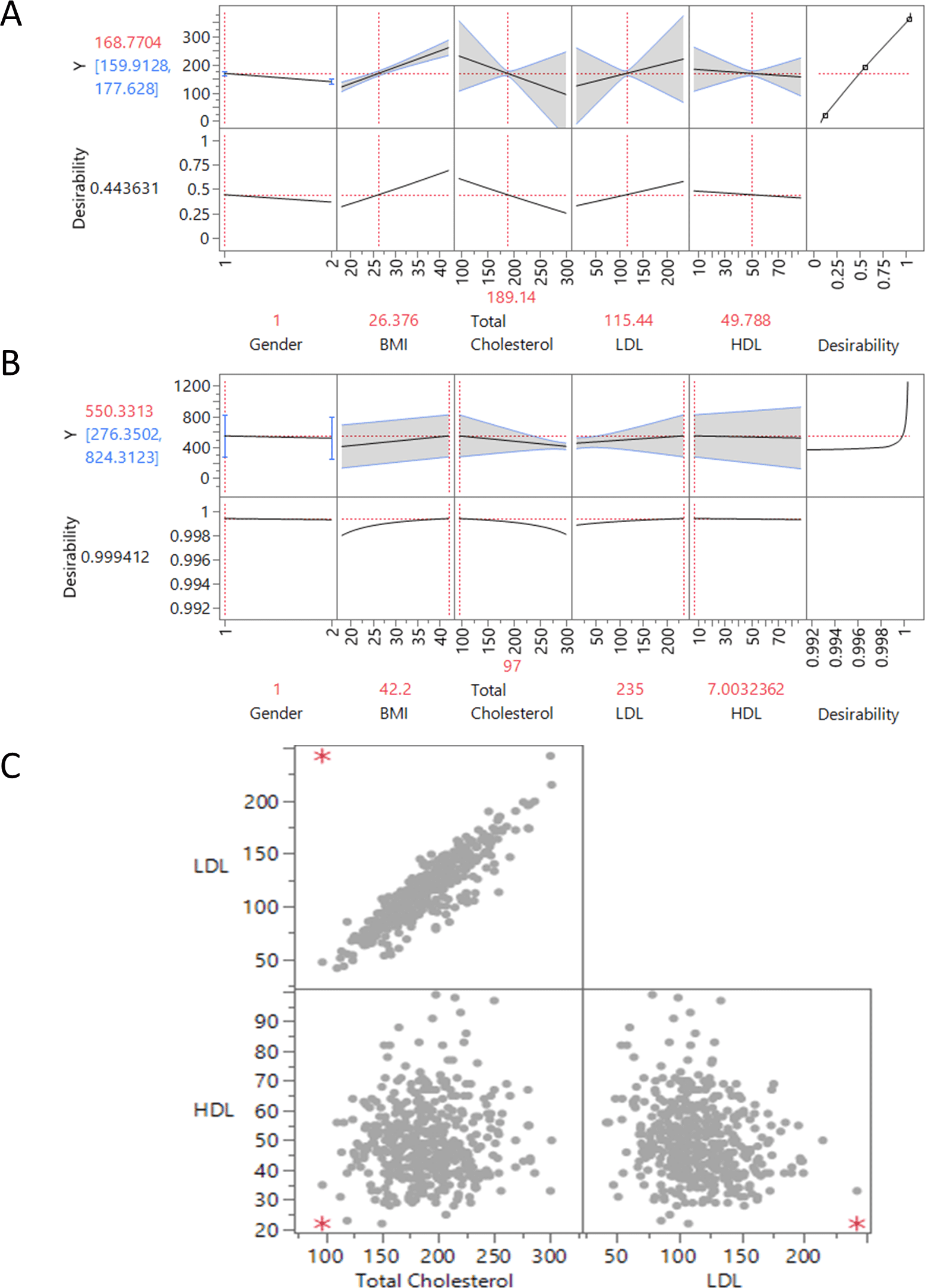} 
	\caption{Example of a prediction profiler for a least squares model fit to the diabetes data discussed in Section 4.1. Predictions are shown for the response variable on the y-axis. The model factors are shown on the x-axis.  Only a subset of the factors are shown for simplicity. 95\% onfidence intervals are shown in grey. The desirability function has been selected to increase linearly with the response. A) Continuous factors are initialized to their mean values in the training data and categorical factors are initialized to the highest frequency level.  B) Factor settings when the response variable (Y) is maximized. C) Scatterplot for three factor variables.  The training data (grey) and the prediction point where the response is maximized (red). Since total cholesterol is a function of the sum of HDL and LDL, the high LDL and low total cholesterol observed in the prediction point is not physically realizable.}\label{fig:profile_example}
\end{figure}

When predictive models contain many factors, the prediction surface of the model is high dimensional and can be challenging to visualize. This makes it difficult to explore how predictions will change in response to changes in factor settings.  A graphical tool that addresses this problem is the prediction profiler \citep{jones1991interactive, jones2021prediction}, hereafter referred to as the profiler, which allows the user to visualize predictions over a high dimensional factor space. \textbf{Figure \ref{fig:profile_example}A} shows an example of a profiler in the JMP software.  The diabetes data and the least squares model shown in this figure will be discussed in more detail later in the results section.  For each model factor, a cross sectional view, or profiler trace, of the prediction surface is shown. A profiler trace shows how predictions change as a function of a model factor, with all other factors set at fixed values. Sliders for each factor allow users to explore how the cross-sectional views change as the factor settings change. Users can also set desirability functions, or optimization goals, for each response in the profiler (e.g., maximize, minimize, or match target) and perform optimization.  Models with multiple responses can be accommodated in the profiler, where a row of plots corresponds to each response.


The model factor space in the profiler is set by default to have rectangular boundary constraints for continuous factors that are given by the range of the data.  Users may also add mixture constraints or other linear constraints. The model factor space for categorical factors is the discrete grid of factor levels.  Defining the factor space in this way does not prevent the user from exploring prediction points that should be considered extrapolation. The particular class of extrapolations that we focus on in this paper are violations of the correlation structure of the data. \textbf{Figure \ref{fig:profile_example}B-C} provides an example were the predicted response has been maximized and there are strong violations of the factor correlation structure. Extrapolation can be dangerous for a number of reasons:

\begin{itemize}
    \item Predictions rely on model assumptions that may become more invalid the further the prediction is from the original data.
    \item Prediction uncertainty increases the further the prediction is from the original data.
    \item Extrapolations may involve factor combinations that cannot be realized physically. For example, in \textbf{Figure \ref{fig:profile_example}C}, total cholesterol is a function of the sum of HDL and LDL. It is not possible to observe high LDL and low total cholesterol as observed in the prediction point obtained in the profiler.
\end{itemize}

\noindent In earlier iterations of the profiler, it was easy to achieve extrapolated predictions and difficult to know that the predictions were extrapolated. One could plot the predictions on accompanying plots, such as a scatterplot matrix of the factor variables, but this process was often tedious, especially when the number of factors was large. The goal of this paper is to provide a method for determining if predictions are extrapolated directly within the profiler. 

There are a wide variety of methods used to assess whether a prediction point should be considered extrapolation.  The fields that have developed the methodology often have both high dimensional and highly correlated factors.  For example, in cheminformatics \citep{eriksson2003methods, netzeva2005current} and chemometrics \citep{rousseeuw2006robustness}, the chemical features are often highly correlated.  In these areas it is common for new data to be collected from subpopulations that are not observed in the training data, so it is often necessary to screen out any data that should be considered extrapolation prior to performing prediction. Similar methodology can also be found in other fields, such as ecology \citep{mesgaran2014here, bartley2019identifying}.  

See \cite{Mathea2016} for a review of the popular metrics used for defining extrapolation in predictive models.  Some examples are leverage, Hotelling's $T^2$, convex hull based methods, nearest neighbor based estimation of local density, kernel density estimation and 1-class SVM.  We utilize two metrics in this study: the well known leverage metric, which is specific to least squares linear models, and a novel regularized Hotelling's $T^2$ metric, which is designed for machine learning models in general. The basic workflow for our method is:

\begin{itemize}
    \item Compute an extrapolation measure for all observations in the training data.
    \item Use the extrapolation measures to establish a threshold, beyond which prediction points are classified as extrapolation.
    \item Use the classifications to either restrict profiler traces so that extrapolations are inaccessible to the user, or warn the user when they move into an extrapolated region. 
    \item If a user requests optimization of a desirability function, perform optimization subject to the extrapolation constraint.
\end{itemize}


The structure of the paper is as follows. Section 2 presents details on the extrapolation control workflow: the distance metrics used, the procedure for obtaining thresholds for the determination of extrapolation, and the genetic algorithm used for optimization. Section 3 evaluates the workflow using simulation studies.  Section 4.1 demonstrates the extrapolation control workflow for least squares linear models on a real world example. Section 4.2 demonstrates the extrapolation control workflow for general machine learning models on a real world example.

\section{Methods}
\label{sec:meth}

\subsection{Least squares models} \label{sec:LS}

For least squares linear models we chose leverage, a well known metric used to identify outliers in linear models. Let $\M{H} = \M{X}(\M{X}^T\M{X})^{-1}\M{X}^T$ be the hat matrix for a linear model with $n \times p$ design matrix $\M{X}$.  $h_{ii}$, which is the $i$th diagonal element of $\M{H}$, is the leverage for observation $i$ in the training data. A typical point will have average leverage, which is $\frac{1}{n}\sum_{i= 1}^n h_{ii} = \frac{p}{n}$. The leverage for a new prediction point is computed as $h_{pred} = \V{x}_{pred}^T(\M{X}^T\M{X})^{-1}\V{x}_{pred}$, where $x_{pred}$ is a $p$-dimensional new prediction observation.

One interpretation of $h_{pred}$ is that is the multivariate distance from the center of the training data in the factor space.  Another interpretation is that it is the scaled prediction variance.  That is, as the prediction point moves further away from the center of the data (in a multivariate sense) the uncertainty of the prediction increases.

There are two criteria commonly used in the statistical literature \citep{cook1977detection, bartley2019identifying} for determining if a prediction point with leverage $h_{pred}$ should be considered extrapolation:

\begin{itemize}
\item $h_{pred} > k \cdot max(h_{ii})$, where $k$ is a customizable multiplier. A typical value of $k$ is 1. The $max(h_{ii})$ is the leverage of the furthest point on the convex hull of the training data in factor space.  When $k = 1$, prediction points beyond the threshold are outside of the convex hull.
\item$h_{pred} > l \cdot \frac{p}{n}$, where $l$ is a customizable multiplier. Typical values of $l$ are 2 and 3. Recall that $\frac{p}{n}$ is the average leverage for the training data.
\end{itemize}

\subsection{General machine learning models}

When selecting a generalized extrapolation control metric for models other than linear models, we had a number of criteria.  First, we needed a metric that was computationally efficient for a large number of factors and observations. This was necessary to maintain the interactivity of the profiler traces and to preform optimization efficiently. Next, we wanted to be able to support continuous, categorical and ordinal factors.   We also wanted to utilize observations with missing cells, because some modeling methods include these observations in the fit. We wanted a method that was robust to linear dependencies in the data, these occur when the number of variables is larger than the number of observations, for example. We also wanted something that was easy to automate, without much user input. And finally, we wanted something that was easy to explain to non-statisticians.

We restricted our focus to extrapolation control methods that were unsupervised.   That is, we only flag a prediction point as extrapolation if it is far outside the distribution of the training data in factor space.  We do this so that our extrapolation control metric will generalize to wide variety of machine learning models. This was also necessary to be consistent across model profilers, so that profilers can be linked and models can be ensembled.

We also restrict our consideration to extrapolation control methods that protect against violations of correlation structure in the training data.  This type of extrapolation is the major concern in the use cases we focused on, where popular methodology is based on $T^2$ and square prediction error (SPE) metrics for principal components analysis and partial least squares models \citep{eriksson2003methods, rousseeuw2006robustness}. We discuss future generalizations of this approach to protect against extrapolations between clusters of data in the discussion.

The multivariate distance interpretation of leverage suggested Hotelling’s $T^2$ as a distance metric for general extrapolation control. In fact, there is a monotonic relationship between Hotelling’s $T^2$ and leverage. Since we are no longer in linear models, this metric does not have the same connection to prediction variance.  So instead of relying on the extrapolation thresholds used for linear models, we make distributional assumptions about $T^2$ to determine an upper control limit to be used as the extrapolation threshold.



The formula for Hotelling’s $T^2$ is $T^2 = (x-\bar{x})^T\hat{\Sigma}^{-1}(x-\bar{x})$.  The mean and covariance matrix for the factors are estimated using the training data for the model. If $p < n$ and the factors are multivariate normal (MVN), then $T^2_{pred}$  for a prediction point has an F distribution: $T_{pred}^2 \sim \frac{(n+1)(n-1)p}{n(n-p)} F(p, n-p)$.  However, we wanted the method to generalize to data with complex data types such as a mix of continuous and categorical variables, data sets where $p > n$, and data sets where there are a large fraction of missing values.  Instead of determining the distribution of $T^2_{pred}$ in each of these scenarios, we use a simpler and more conservative control limit that we found 
works well in practice: a 3-sigma control limit that uses the empirical distribution of $T^2$ based on the training data.  The control limit, that we use as our extrapolation threshold, can be calculated by the following equation: $UCL = \bar{T}^2 + 3\hat{\sigma}_{T^2}$ where $\bar{T}^2$ + is the mean of training data $T^2$ and $\hat{\sigma}_{T^2}$ is the standard deviation of the training data $T^2$.

Next, we describe a novel approach to computing the Hotelling's $T^2$, which we refer to as Regularized $T^2$.  One complication with the standard Hotelling’s $T^2$ is that it is undefined when $p > n$, because there are too many parameters in the covariance matrix to estimate with the available data.  To address this we use a regularized covariance matrix estimator developed by  \cite{Schafer2005}.  This estimator was originally developed for the estimation of covariance matrices for high-dimensional genomics data.  The estimator has been shown to produce a more accurate covariance matrix estimate when $p$ is much larger than $n$. The estimator is:

\begin{align*}
\hat{\Sigma} = (1-\hat{\lambda})\hat{\M{U}} + \hat{\lambda}\hat{\M{D}}.
\end{align*}

\noindent This is simply a weighted combination of the full sample covariance matrix ($\hat{\M{U}}$) and a constrained target covariance matrix ($\hat{\M{D}}$).  The target matrix we used is a diagonal matrix, with the sample variances of the factor variables on the diagonal and zeroes on the off-diagonal.  This shrinkage estimator is guaranteed to be positive definite when $p > n$.  This is necessary to compute Hotelling’s $T^2$ when $p > n$. For the $\hat{\lambda}$ weight parameter, Schafer and Strimmer derived an analytical expression that minimizes the mean squared error of the estimator asymptotically.

Schaffer and Strimmer proposed several possible target matrices. The diagonal target matrix $\hat{\M{D}}$ assumes that variables are uncorrelated as the prior.  This works well for a general extrapolation control method, as we do not assume any correlation structure between the variables without prior knowledge of the data. Also, when there is little data to estimate the covariance matrix, the elliptical extrapolation control constraints are expanded by the shrinkage estimator with the diagonal target matrix. This results in a more conservative test statistic for extrapolation control. That is, when there is little data available to estimate the covariances, tests based on Regularized $T^2$ are less likely to label prediction points as extrapolation.  This is sensible, as covariances may be observed by chance when training data is limited.

Another complication that needed to be addressed was how to compute Hotelling’s $T^2$ when there is missing data.  Many predictive modeling methods can utilize observations with missing values for prediction, such as random forests and boosted trees. When the sample size of the training data is small and there is missing data, it would be ideal to do pair-wise deletion instead of row-wise deletion to estimate the covariance matrix. This increases the amount of data that is available to estimate the entries. Let $\hat{\M{U}} = ((u^{kl}))$ be the covariance matrix estimator. The following formulas show the pairwise deletion method:

\begin{equation*}
\begin{gathered}
\bar{x}^k = \frac{\sum_{i=1}^{n} x_i^k \mathbbm{1}(x^k_i \neq NA)}{\mathbbm{1}(x^k_i \neq NA)} \\
u^{kl} = \frac{\sum_{i=1}^{n} (x_i^k - \bar{x}^k)(x_i^l - \bar{x}^l) \mathbbm{1}(x^k_i \neq NA, x^l_i \neq NA)}{\mathbbm{1}(x^k_i \neq NA, x^l_i \neq NA)}
\end{gathered}
\end{equation*}

Previously, the profiler allowed constrained optimization with linear constraints.  Since extrapolation control is a non-linear constraint, the optimization problem is more challenging.  A genetic algorithm has been implemented to perform the optimization.

\section{Simulation Studies}
\label{sec:sim}

\begin{figure}
	\centering 
	\includegraphics[width=.75\linewidth]{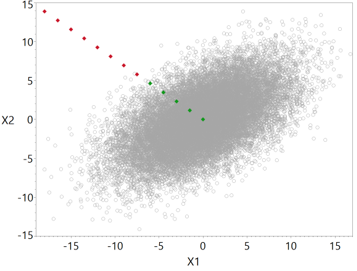} 
	\caption{Example grid of prediction points classified as extrapolation (red) or not extrapolation (green).}\label{fig:5}
\end{figure}

To evaluate the extrapolation control performance of Regularized $T^2$, we performed a simulation study. First, we simulate a factor matrix with a low rank approximation. We do this to evaluate our ability to detect violations of the correlation structure in the data:

\begin{equation*}
\M{X}_{n \times p} =  \M{U}_{n \times r} \M{D}_{r \times p}  + \V{e}_{n \times 1} 
\end{equation*}

\noindent where $r$ is the desired the rank, and each element of $\M{U}$, $\M{D}$, and $\V{e}$ is i.i.d. standard normal (\textbf{Figure \ref{fig:5}}). 


For each simulated data set, we chose the pair of most correlated variables. We then extend a grid of equally spaced points from the center of the data to the corner of the box constraint in the profiler, which is the range of the factors according to the training data. Since the factor variables are multivariate normal (MVN) with known covariance matrix, $T^2$ is $\chi^2$-distributed.	We then use the probability that $T^2_{pred}$ for a grid point is from the same distribution as the data to classify points as extrapolation or not extrapolation:

\begin{equation*}
  \left\{
  \begin{array}{@{}ll@{}}
    \text{``extrapolation"}, & \text{if}\ P(T^2 \geq T^2_{pred}) < .05 \\
    \text{``not extrapolation"}, & \text{if}\ P(T^2 \geq T^2_{pred}) \geq .05
  \end{array}\right.
\end{equation*}

\begin{figure}
	\centering 
	\includegraphics[width=.95\linewidth]{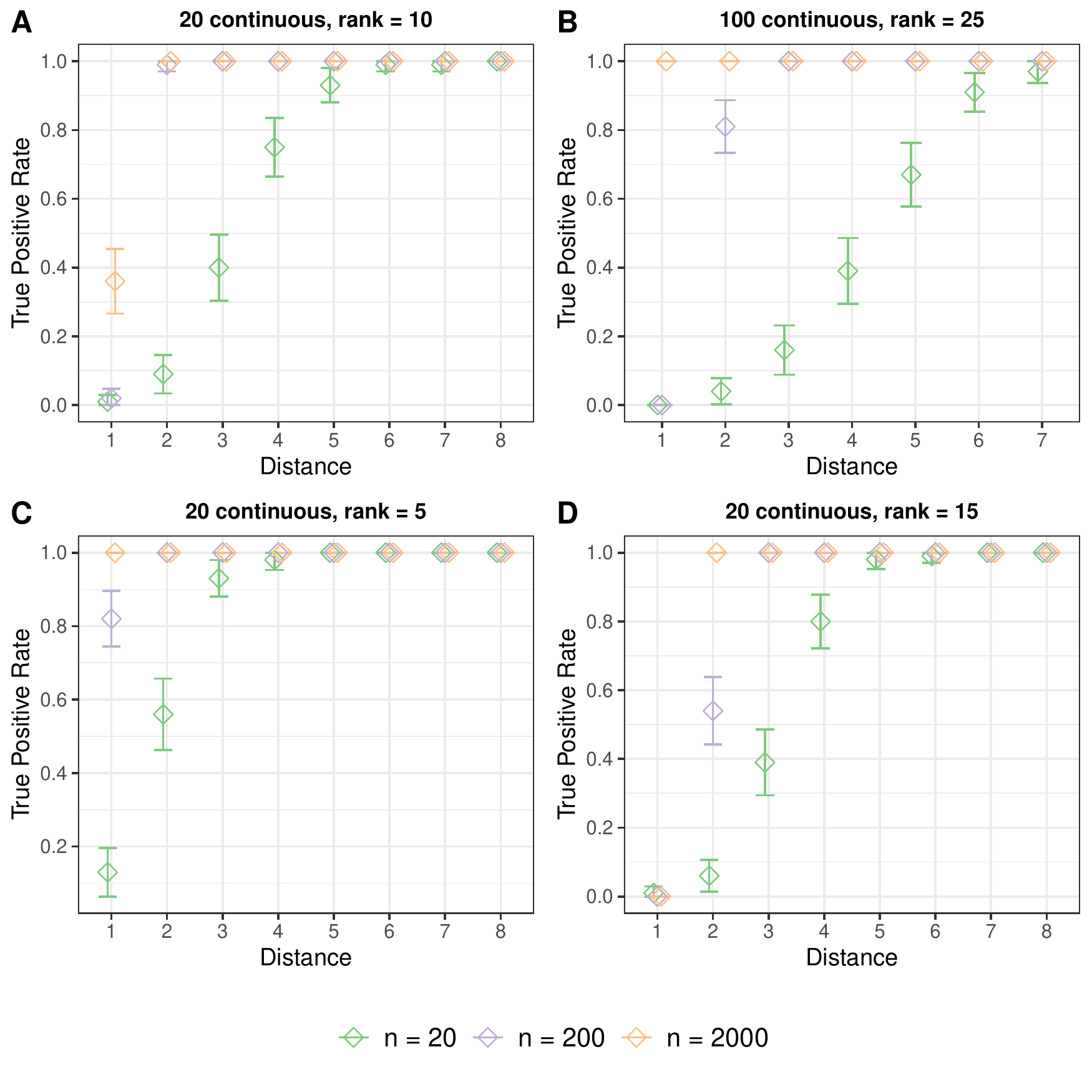} 
	\caption{True positive rates of classification of extrapolated points in simulated data consisting of entirely continuous factors. Various sample sizes, number of factors and rank of factor matrix are shown. The new prediction points are ranked by the extent of extrapolation.  100 simulation replicates were performed. Error bars show 95\% confidence intervals.}\label{fig:6}
\end{figure}

To evaluate Regularized $T^2$ extrapolation control in terms of both false positive rate (FPR) and true positive rate (TPR), we simulated data with increasing sample sizes using the true MVN distribution.  We then used the simulated data to compute a 3-sigma limit, and we determined how well we were able to classify the new prediction points as extrapolated or not. For all of the scenarios shown in \textbf{Figure \ref{fig:6}} the FPR is less than .05, so only the TPR for extrapolated points are shown. On the x-axis, the new prediction points along the grid are ranked by the extent of extrapolation. Along the panels, the number of variables and the rank of the factor matrix is varied. \textbf{Figure \ref{fig:6}A} shows a case when there are 20 factor variables with rank 10.  Because the threshold is conservative, notice that the TPR is low when the extrapolation is mild.   When there is limited training data, the Regularized $T^2$ is less likely to label predictions as extrapolation.  This is desirable because covariances may be observed by chance when training data is limited. However, the TPR does improve the larger the sample size. 

Also note that the TPRs are high in all cases when $p = 20$ and $n = 20$. This is when $p = n$, and when the rank of the covariance matrix, $r_{cov}$, is equal to $n-1$.  In these cases, Hotelling's $T^2$ is undefined, since the covariance is not full rank.  If a generalized inverse, such as the Moore-Penrose inverse, is used to invert the covariance matrix, then the Hotelling's $T^2$ for all training set observations will be a constant, $(2r_{cov})^{1/2}$. Our 3-sigma control limit for Hotelling's $T^2$ will be too large are too small depending on $r_{cov}$.  \textbf{Figure \ref{fig:S0}} shows a case when the Hotelling's $T^2$ control limit is too small, leading to extrapoltion control with high FPRs, and a case when the limit is too large, leading to extrapolation control with low TPRs.  Regularized $T^2$ achieves improved extrapolation control in both cases.

When $p = 100$ and $r = 25$ (\textbf{Figure \ref{fig:6}B}), the TPRs are low when $p$ is much larger than $n$.  This is a hard problem for distance based methods due to the curse of dimensionality, but the TPRs do improve as the sample size increases. \textbf{Figure \ref{fig:6}C-D} shows how TPR is affected by varying rank with fixed p.  The TPR is better when the intrinsic dimensionality of the data is low. 

\begin{figure}
	\centering 
	\includegraphics[width=.85\linewidth]{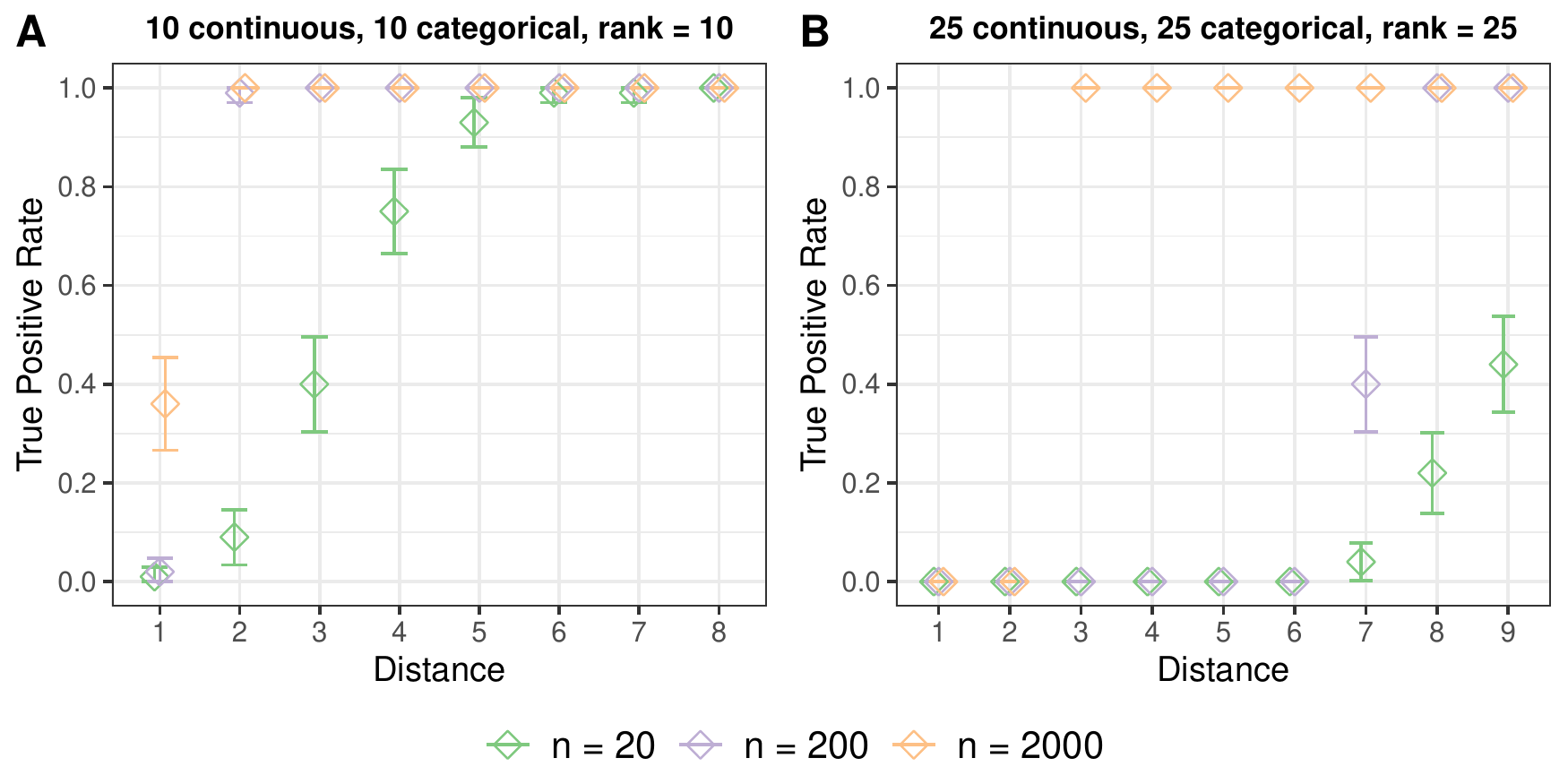} 
	\caption{True positive rates of classification of extrapolated points in simulated data consisting of a mix of categorical and continuous factors. The new prediction points are ranked by the extent of extrapolation.  100 simulation replicates were performed. Error bars show 95\% confidence intervals.}\label{fig:7}
\end{figure}

We also evaluated Regularized $T^2$ extrapolation control on simulated data with a mix of continuous and categorical variables (\textbf{Figure \ref{fig:7}}). We simulated an entirely continuous factor matrix using the same procedure as previously described. We then selected a number of variables ($p_{cat}$) to transform into categorical variables by randomly selecting the number of categories from 2 to 4 for each variable and using equally spaced quantiles to discretize each variable.  \textbf{Figure \ref{fig:7}} shows two scenarios where the number of factor variables are 20 and 50. Half of the variables are categorical, and the rank of the factor matrix is half of the number of variables.	When $p = 20$, our method is able to obtain high TPRs for points that are at least moderately extrapolated, but the problem becomes more challenging when $p >> n$. 


\section{Applications}
\label{sec:verify}

\subsection{Diabetes data: Least square model example}

We demonstrate extrapolation control for least squares models using a diabetes data set originally analyzed in \cite{efron2004least}. In the data, ten factor variables have been measured on 422 diabetes patients.  The variables include age, gender, body mass index, average blood pressure, and six blood serum measurements. Note that this data has mix of continuous and categorical factors, as gender is categorical. The response (Y) is a continuous measure of diabetes progression. 133 observations were randomly held out as a validation set.  A least squares model was fit to the remaining observations (Training $R^2$ = .54, Validation $R^2$ = .44).

\begin{figure}
	\centering 
	\includegraphics[width=.55\linewidth]{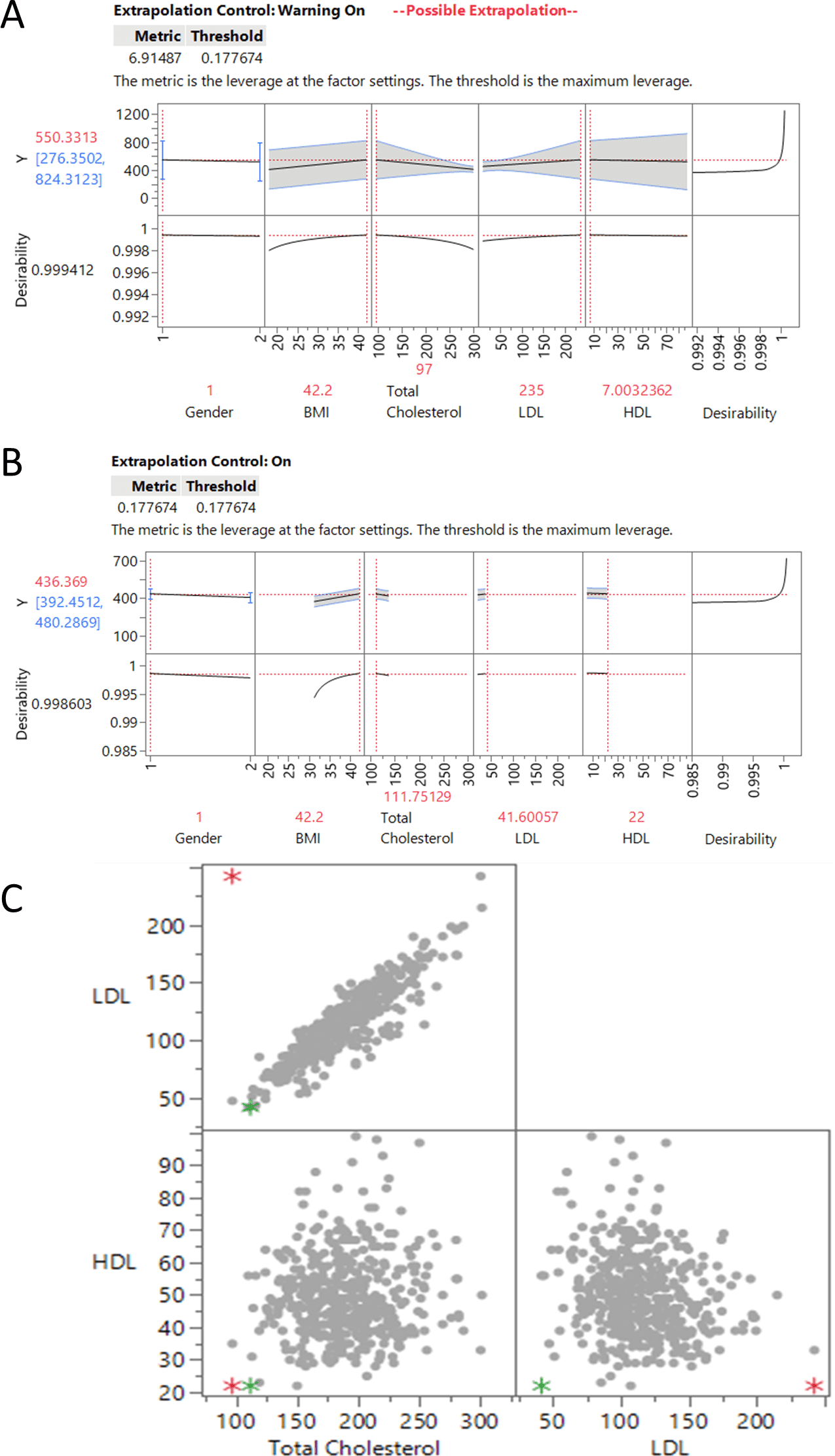} 
	\caption{Extrapolation control in a profiler for a least squares linear model fit to the diabetes data. (A) Profiler traces with extrapolation control set to ``warn" and factors at the settings found to maximize the response in Figure \ref{fig:profile_example}. Leverage is far above the extrapolation threshold, which triggers the extrapolation warning. (B) Profiler traces with extrapolation control set to ``constrain". Factors are set to the prediction point maximizing the response, subject to the extrapolation control constraint. (C) The prediction point without extrapolation control (red) violates the correlation structure of LDL and Total Cholesterol, while the extrapolation control prediction (green) does not.}\label{fig:LS_profile}
\end{figure}

The extrapolation control feature in the prediction profiler has two settings: \textit{warn} and \textit{constrain}. The \textit{warn} setting warns users when factors are set to values that are beyond the extrapolation threshold. \textbf{Figure \ref{fig:LS_profile}A} shows profiler traces for six out of the ten factors for simplicity.  The factors are set to the values found to maximize the response in \textbf{Figure \ref{fig:profile_example}}.  The leverage for this prediction point (8.62) is far above the maximum leverage for the training data (.18).  This has triggered the extrapolation warning.

The \textit{warn} setting is a permissive mode that only warns the user that prediction points may be suspect. A stricter option is \textit{constrain}.  This restricts the profiler traces such that extrapolated regions of the factor space are inaccessible (\textbf{Figure \ref{fig:LS_profile}B}). This is similar to how the profile traces are restricted in the presence of linear constraints, except the leverage constraint is a non-linear, elliptical constraint.



\begin{figure}[H]
	\centering 
	\includegraphics[width=.95\linewidth]{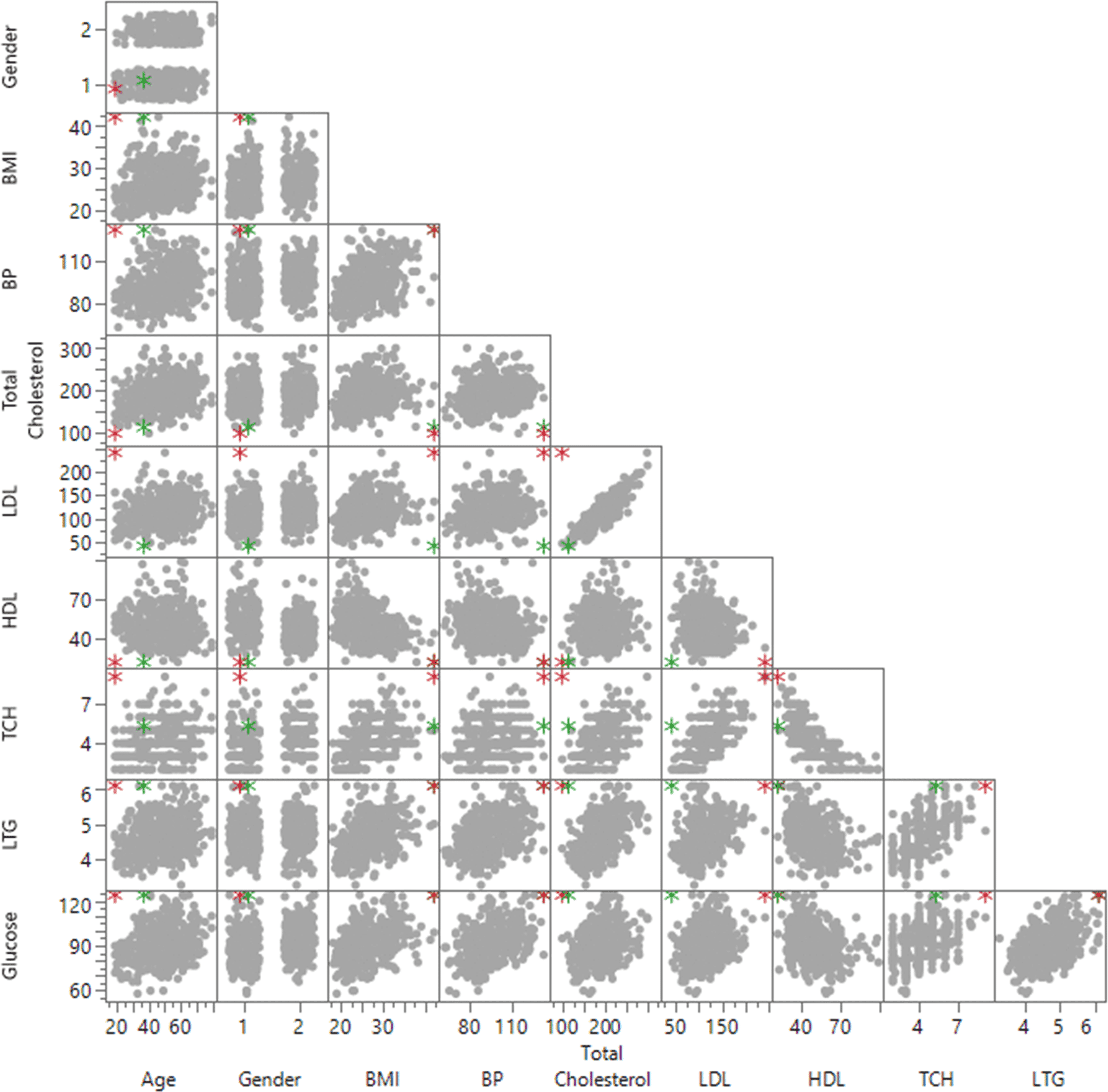} 
	\caption{Scatterplot matrix showing the optimal prediction point without extrapolation control (red) and with extrapolation control (green) for the diabetes data.}\label{fig:LS_scatter}
\end{figure}

Extrapolation control is particularly important when optimizing desirability or using a model to match a target value.  To optimize with extrapolation control, we simply set extrapolation control to \textit{warn} or \textit{constrain} prior to optimization.  \textbf{Figure \ref{fig:LS_profile}B} shows the factor settings when desirability has been optimized with extrapolation control. The unconstrained solution shows a large improvement in shrinkage, but the predicted value is far outside of the range observed in the training data. \textbf{Figure \ref{fig:LS_profile}C} shows how the solution has been constrained to remain within the correlation structure of HDL and Total Cholesterol.  \textbf{Figure \ref{fig:LS_scatter}} shows that extrapolation has been controlled for many other pairs of variables.


\subsection{Metallurgy data: Machine learning model example}

We demonstrate generalized extrapolation control using a manufacturing data set. The data are from a company that manufactures steel drive shafts for the automotive industry using powder metallurgy. The surface of the parts must be free of pitting, or porosity, and if any pitting is observed the entire production run must be scrapped.  The company wishes to find the conditions that minimize the chance of failure.

Key process variables and features of the parts have been recorded for 6,253 production runs.    The manufacturing process begins with a metal powder. Pressure is applied to compact the powder into a desired part shape, and one key process variable is the amount of compaction pressure. Next, a sintering furnace heats the metal particles so that they bind together, which strengthens the metal part. Another process variable is sintering temperature, which is the average temperature in the sintering furnace. Two response variables have been collected.  One is surface condition, a binary response indicating whether the product was a failure or not, and another is shrinkage, a continuous response that is associated with surface condition, where large values of shrinkage are associated with a high failure rate.

We trained a boosted neural network to predict both response variables as a function of the process variables.  Weak learners were desirable for the boosting iterations, so at each iteration we trained a neural network with one hidden layer and three neurons.  Each neuron had a TanH activation function. 20 boosting iterations were used. To validate our model, we held out approximately 25\% (1,563 observations) as a validation set. We performed stratified sampling with both responses as stratification variables.  

Next, we assess whether the model had reasonable prediction performance for the response variables. \textbf{Figures \ref{fig:S1}} shows performance for the binary surface condition response. Because the classes were highly imbalanced (the frequency of failures was .045 in the entire dataset), precision and recall were used as the primary metrics for model evaluation. On the training set, a threshold of .05 on the predicted probability of failure was found to achieve a balance between recall (.75) and precision (.14).  With this threshold, the model also achieved reasonable performance on the validation set (recall = .69, precision = .13). \textbf{Figures \ref{fig:S2}} shows the model obtained good prediction performance for continuous shrinkage variable, as well (Training $R^2$ = .84, Validation $R^2$ = .83).

\begin{figure}
	\centering 
	\includegraphics[width=.95\linewidth]{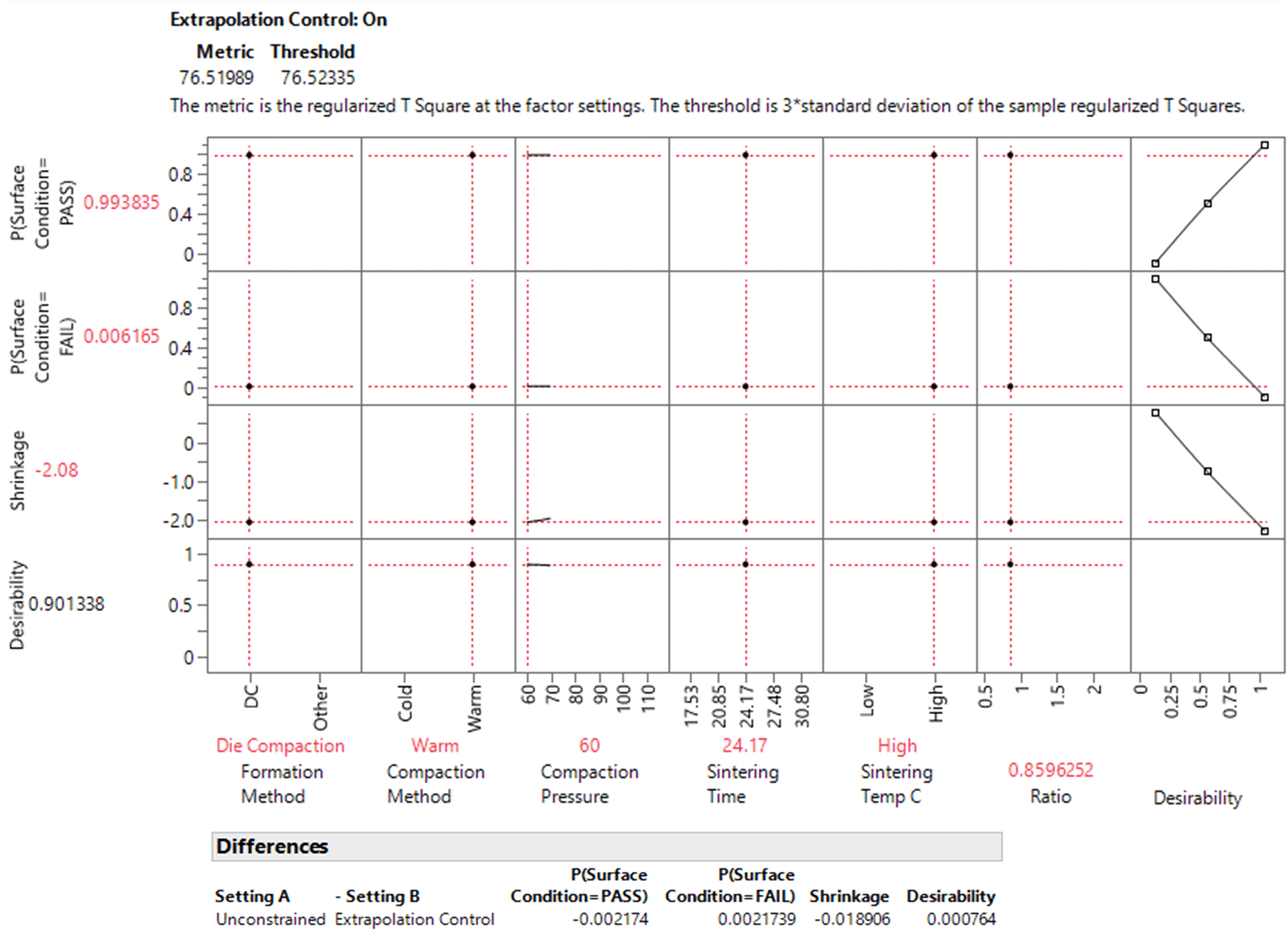} 
	\caption{The optimal prediction point in the boosted neural network profiler with extrapolation control. Since factor settings that minimize failure rate are desired, we maximize the probability of passing, minimize the probability of failure, and minimize shrinkage.  There is little difference between the desirabilities for the extrapolation control optimal prediction point and the unconstrained optimal prediction point.}\label{fig:8}
\end{figure}

\begin{figure}
	\centering 
	\includegraphics[width=.95\linewidth]{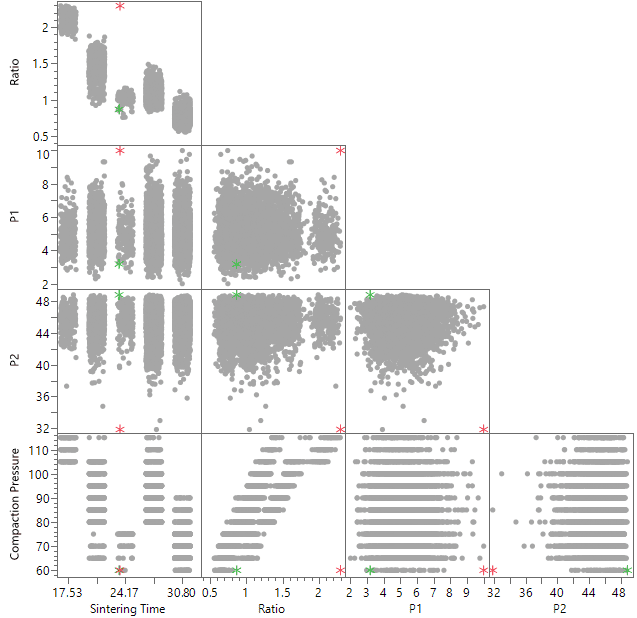} 
	\caption{Scatterplot matrix showing the optimal prediction point without extrapolation control (red) and with extrapolation control (green) according to the boosted neural network trained on the metallurgy data without missing values.  Optimization with extrapolation control obtains a less extrapolated prediction point in comparison to the unconstrained solution. A subset of factors with large violation of correlation structure are shown. See Figure \ref{fig:S3} for the full set of variables.}\label{fig:9}
\end{figure}

Having established that the model obtains acceptable prediction performance, we proceed to finding the factor settings that minimize the chance of failure.  According to surface condition, we maximize the probability of pass and minimize the probability of fail. We also minimize shrinkage. The overall desirability that is maximized for all responses is defined as the geometric mean of the desirability functions for the individual responses. \textbf{Figure \ref{fig:8}} shows the optimized factor settings with extrapolation control, compared to the unconstrained solution. At the top of \textbf{Figure \ref{fig:8}} are the Regularized $T^2$ value at the optimized prediction point with extrapolation control turned on.  Note that it is below the 3-sigma control limit.  \textbf{Figure \ref{fig:9}} shows how the unconstrained solution violates the correlation structure of many of the process variables, and how extrapolation control option corrects this. Note that only a subset of the process variables with the largest extrapolations for the unconstrained solution are shown (\textbf{Figure \ref{fig:S3}} shows the full set of process variables). At the bottom of the \textbf{Figure \ref{fig:8}} is the difference in the desirability between the extrapolation control solution and the unconstrained solution.  The desirabilities are practically equivalent, and the prediction point with extrapolation control obtains much more reasonable factor settings.

Analyzing the same data set, we demonstrate how Regularized $T^2$ can be used when there are missing values. We introduce at random missing values to 50 percent of the entries in the factor matrix.  The boosted neural network with the same parameters was trained with the informative missing values option turned on.  This option mean imputes the missing values and augments the model factors with additional indicator variables, indicating the presence and absence of missing values for each factor.  This was necessary to train a neural model that utilized observations with missing values. By default, neural network models in JMP will remove all observations with missing values prior to training. \textbf{Figures \ref{fig:S4} and \ref{fig:S5}} demonstrate that the model obtained reasonable prediction performance, although performance did detoriate due to the large amount of missing values.  \textbf{Figures \ref{fig:S6}} shows that optimization with extrapolation control obtains a less extrapolated prediction point in comparison to the unconstrained solution.



\pagebreak

\bibliographystyle{Chicago}

\bibliography{bibliography}

\pagebreak

  \bigskip
  \bigskip
  \bigskip
  \begin{center}
    {\LARGE\bf SUPPLEMENTAL MATERIALS}
\end{center}
  \medskip
  
\pagebreak

\appendix
\setcounter{figure}{0}
\renewcommand\thefigure{S.\arabic{figure}}   

\begin{figure}
	\centering 
	\includegraphics[width=1\linewidth]{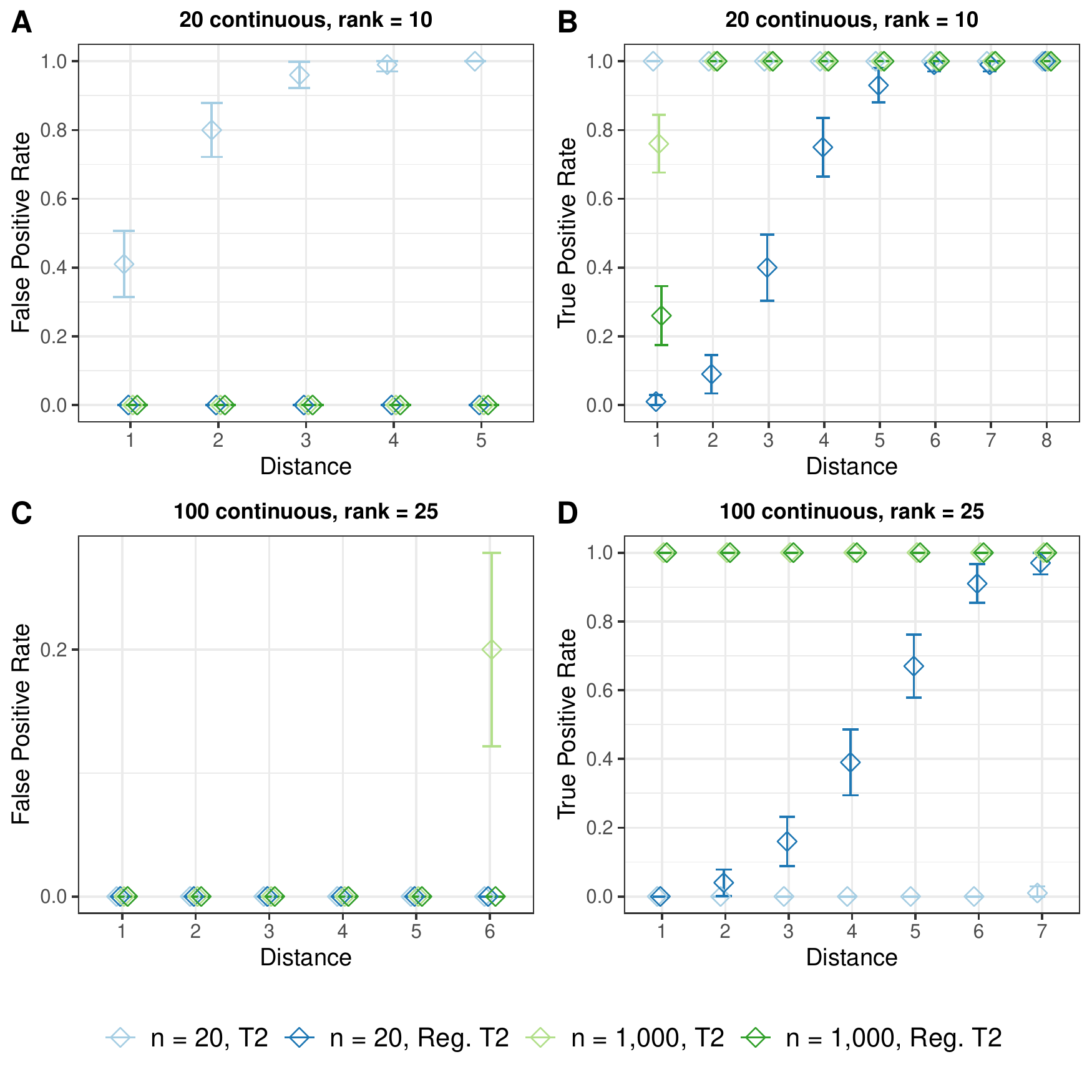} 
	\caption{False positive rates (A, C) and true positive rates (B, D) for classification of extrapolation status of new prediction points using simulated data consisting of entirely continuous factors. The prediction points are ranked by the extent of extrapolation.  100 simulation replicates were performed. Error bars show 95\% confidence intervals. (A-B) the Hotelling's $T^2$ extrapolation threshold is too small, leading to extrapolation control with high FPRs. (C-D) the Hotelling's $T^2$ extrapolation threshold is too large, leading to extrapolation control with low TPRs.  Regularized $T^2$ achieves improved extrapolation control in both cases.}\label{fig:S0}
\end{figure}

\begin{figure}
	\centering 
	\includegraphics[width=1\linewidth]{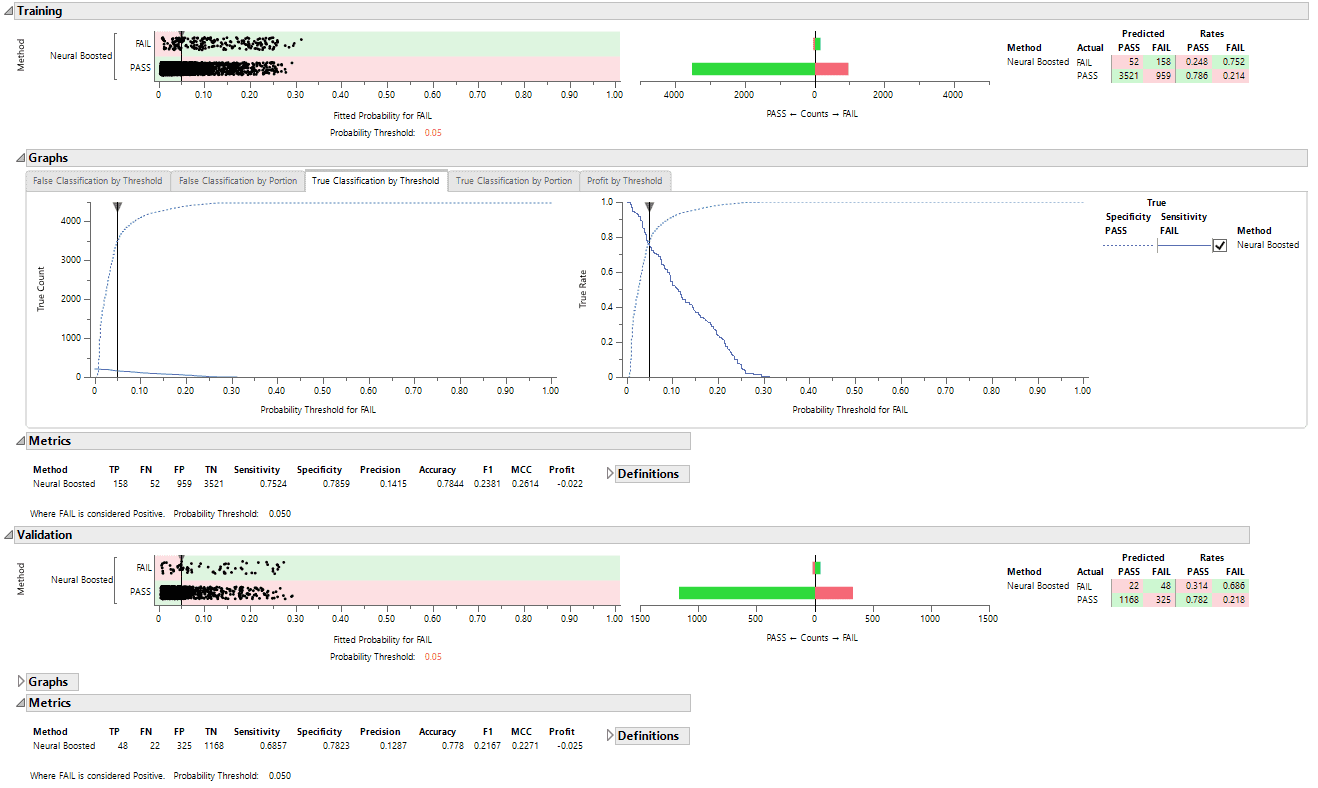} 
	\caption{Classification performance for surface condition according to the boosted neural network trained on the metallurgy data without missing values.  The optimal probability threshold for failure status was selected to be .05 using the training data.  This threshold obtained a balance between precision and recall (sensitivity) in both the training and validation set.}\label{fig:S1}
\end{figure}

\begin{figure}
	\centering 
	\includegraphics[width=1\linewidth]{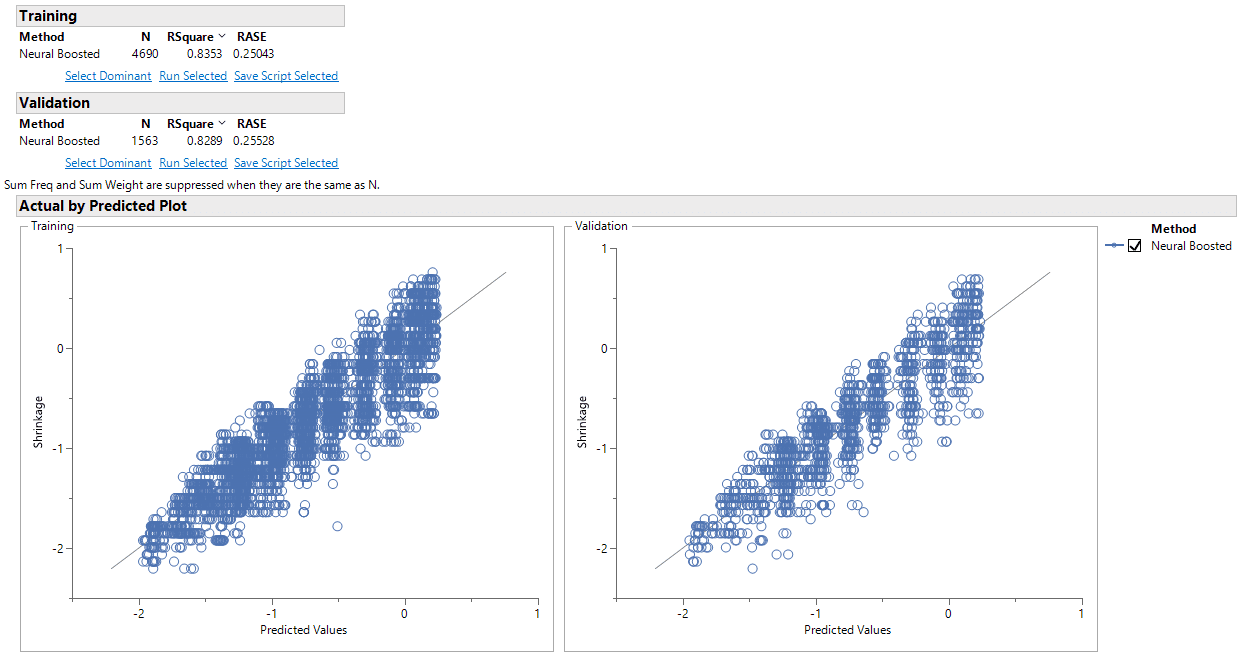} 
	\caption{Prediction performance for shrinkage according to the boosted neural network trained on the metallurgy data without missing values.}\label{fig:S2}
\end{figure}

\begin{figure}
	\centering 
	\includegraphics[width=1\linewidth]{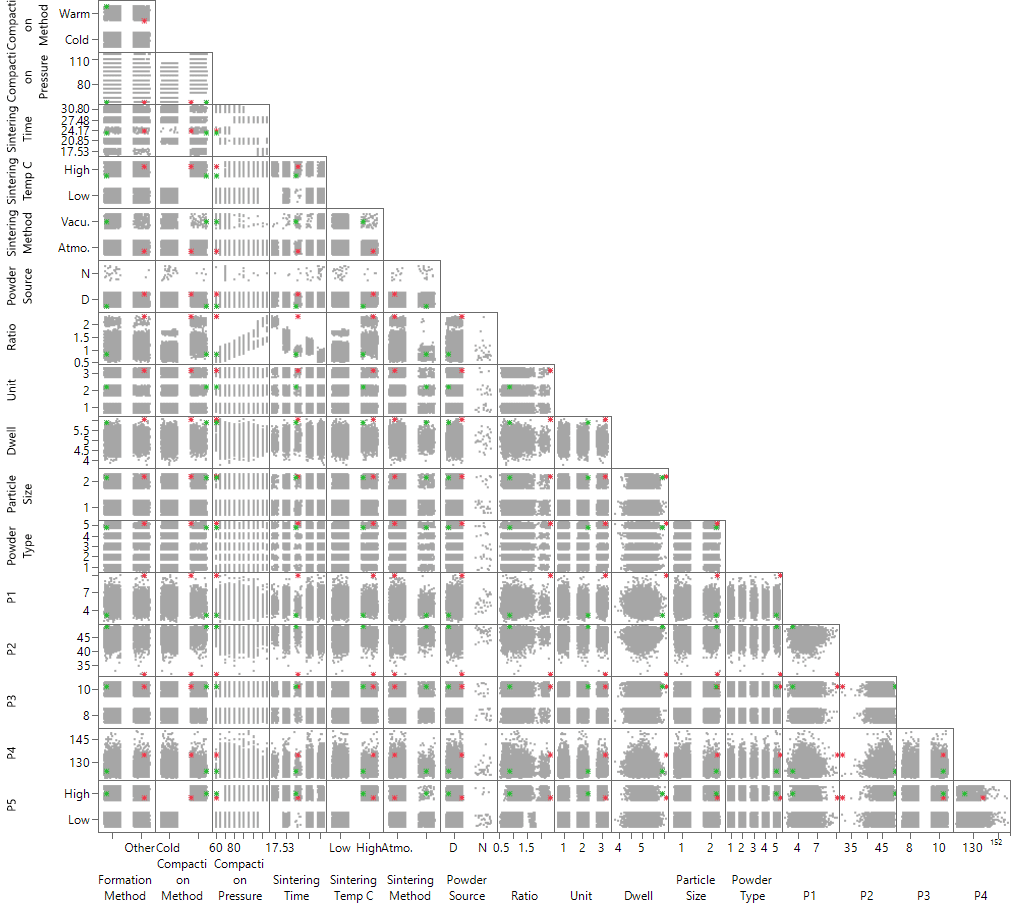} 
	\caption{Scatterplot matrix showing the optimal prediction point without extrapolation control (red) and with extrapolation control (green) according to the boosted neural network trained on the metallurgy data without missing values.  Optimization with extrapolation control obtains a less extrapolated prediction point in comparison to the unconstrained solution.}\label{fig:S3}
\end{figure}

\begin{figure}
	\centering 
	\includegraphics[width=1\linewidth]{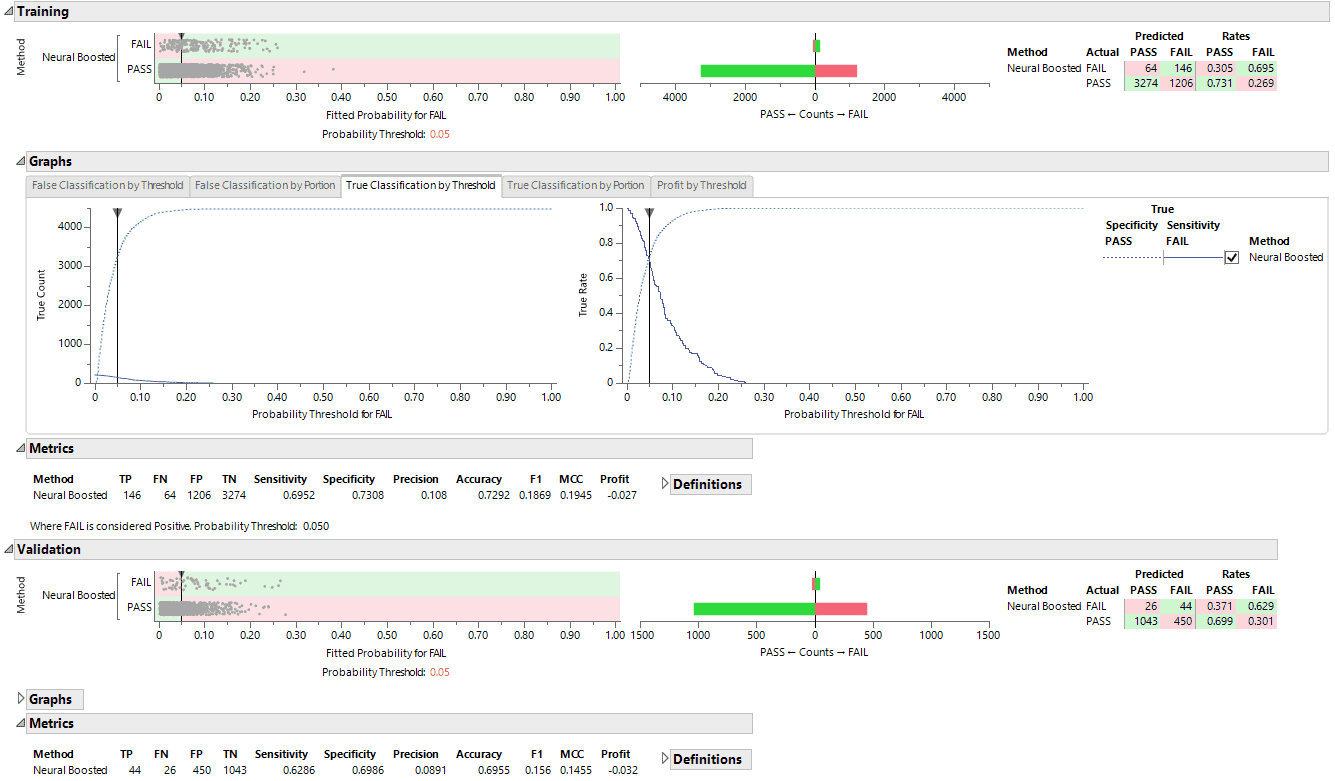} 
	\caption{Classification performance for surface condition according to the boosted neural network trained on the metallurgy data with 50\% missing values.  The optimal probability threshold for failure status was selected to be .05 using the training data.  This threshold obtained a balance between precision and recall (sensitivity) in both the training and validation set.}\label{fig:S4}
\end{figure}

\begin{figure}
	\centering 
	\includegraphics[width=1\linewidth]{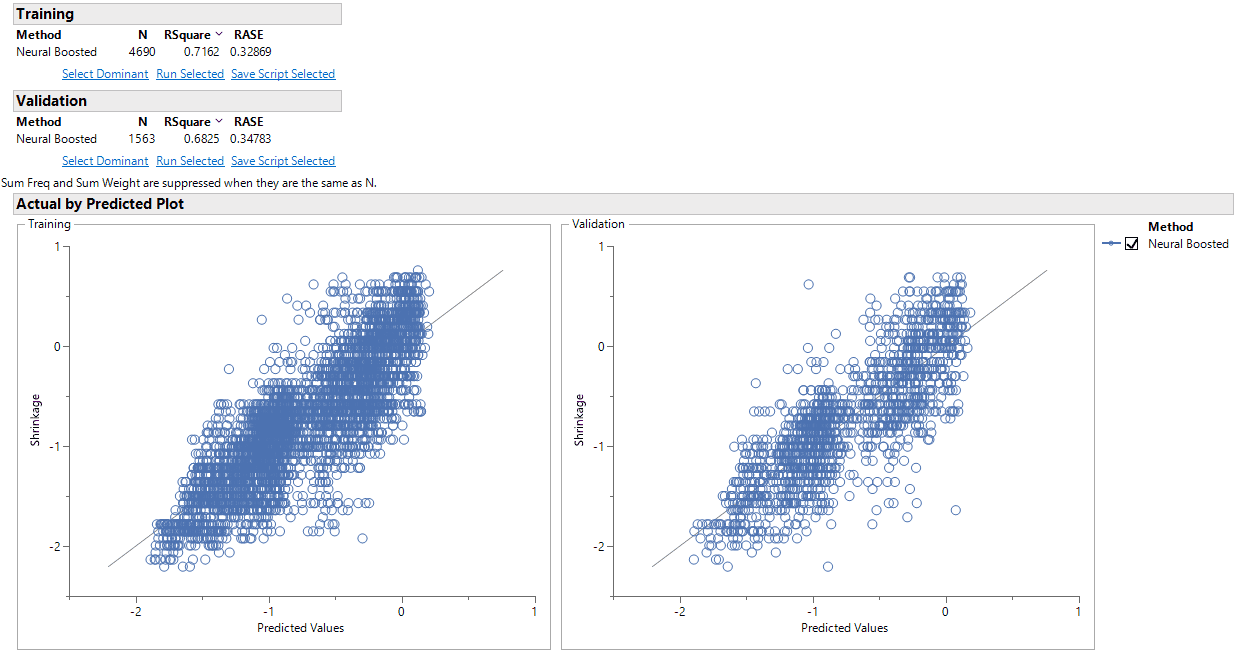} 
	\caption{Prediction performance for shrinkage according to the boosted neural network trained on the metallurgy data with 50\%  missing values.}\label{fig:S5}
\end{figure}

\begin{figure}
	\centering 
	\includegraphics[width=1\linewidth]{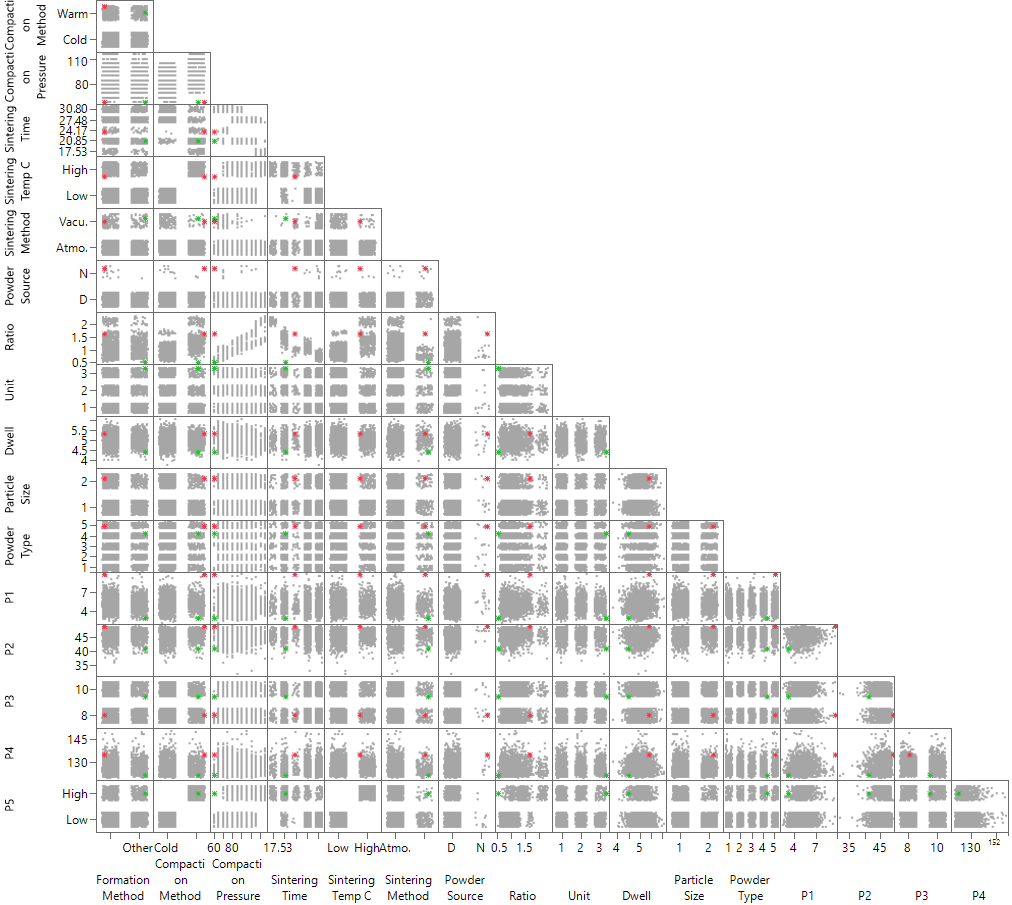} 
	\caption{Scatterplot matrix showing the optimal prediction point without extrapolation control (red) and with extrapolation control (green) according to the boosted neural network trained on the metallurgy data with 50\% missing values.  Optimization with extrapolation control obtains a less extrapolated prediction point in comparison to the unconstrained solution.}\label{fig:S6}
\end{figure}

\end{document}